# Brain-Computer-Interface controlled robot via RaspberryPi and PiEEG.

Ildar Rakhmatulin* – Ph.D. Electronic Researcher
Sebastian Völkl - Brain-Computer-Interface Developer

**Abstract**
This paper presents Open-source software and a developed shield board for the Raspberry Pi family of single-board computers that can be used to read EEG signals. We have described the mechanism for reading EEG signals and decomposing them into a Fourier series and provided examples of controlling LEDs and a toy robot by blinking. Finally, we discussed the prospects of the brain-computer interface for the near future and considered various methods for controlling external mechanical objects using real-time EEG signals.

**Abbreviation**

| | |
|---|---|
| BCI | Brain-computer interface |
| EEG | Electroencephalogram |
| SBC | Single-board computer |
| ADC | Analog-digital converter |

**Introduction**
When the term BCI is mentioned, many people immediately associate with controlling objects using the power of thought. Now, neuroscience in non-invasive EEG measurement is only getting there. Still, each step brings us closer to that goal and inspires a new generation of scientists and engineers to contribute to this field of science. We have machine learning, which entered our lives only a few years ago, and more than enough computing power to look for correlations in EEG signals. The only weakness will be the availability of the dataset. Therefore, we hope that the existence of a device at a low price will allow us to take a step towards solving this problem.
Reading EEG signals, despite its apparent simplicity - measuring microvolts from the scalp through electrodes with a high-precision ADC - is associated with various scientific fields. It involves reading EEG signals [1, 2022], processing EEG signals [2, 2021], selecting features, and finally using the signals for various purposes. Moreover, blinks or chews are unwanted artifacts that introduce harmful distortions into EEG signals, and many works are devoted to combating these artifacts [3, 2022; 4, 2022]. At the same time, however, these artifacts are still commonly used in applied tasks, such as blink control of external objects. Lin et al. [5, 2010] successfully controlled an electric wheelchair by blinking through a brain-computer interface. Huang et al. [6, 2019] developed an application to control the integrated system of a robotic wheelchair by blinking and BCI.
Our board is designed to familiarize everyone with the world of EEG, including those not directly related to the field of neurology. So, our mission is to lower the threshold of technical knowledge to get started with BCI. Our goal was not to compete with the previously described papers and show that our control method has the best performance. Considering all the shortcomings of this method (competition with eye-tracking, low efficiency, limited freedom of action, and complexity in administration), we believe this method is an excellent example to demonstrate the capabilities of BCIs.

**1. Technical details**

The technical points of the developed device PIEEG were presented in the article [7, 2022]. Therefore, in this paper, we only limit ourselves to the main technical details of the device in table 1.

Table 1. PIEEG characteristic

|    | Description | Characteristics |
|----|-------------|-----------------|
| 1  | Channels | 8 channels for connecting electrodes, 1 channel for connecting a reference, and 1 channel for connecting a bias signal, with common-mode noise suppression |
| 2  | Protocol | Data transfer via SPI with a frequency from 250 SPS to 16 kSPS |
| 3  | Resolution | 24 bits |
| 4  | Gain signal | 1, 2, 4, 6, 8, 12, 24 |
| 5  | Impedance | Control connections |
| 6  | Common-Mode Rejection Ratio CMRR | 120 |
| 7  | Internal noise | 0.4 µV |
| 8  | External noise | 0.8 µV |
| 9  | Noise/Ration SNR | 130 dB |
| 10 | Indication status | Of power and live bits for ADS1299 |
| 11 | User switch | 3 |

Fig. 1 schematically shows the use of the developed device for robotic objects control.

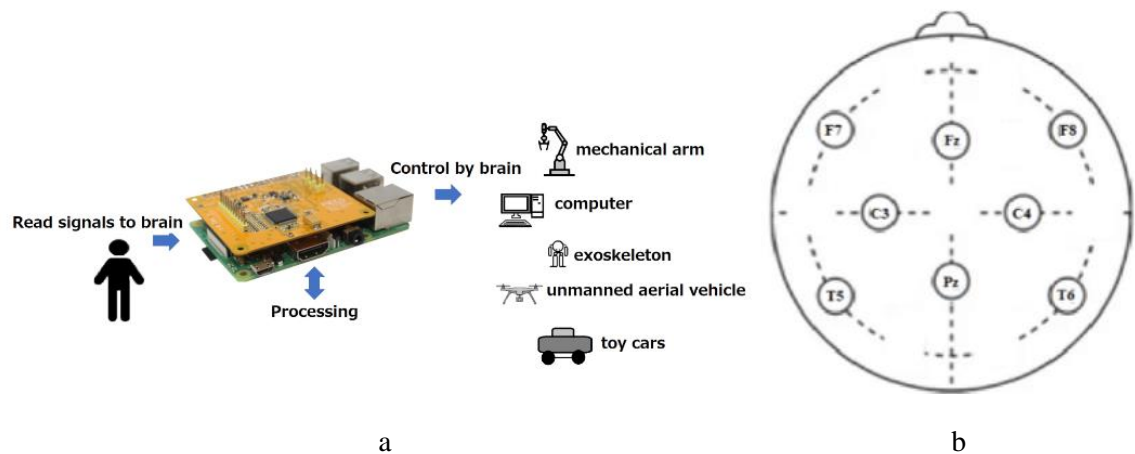

a　　　　　　　　　　　　　　　　　　b

Fig. 1.a - Schematic representation of the use of PIEEG, b -8 electrodes connected by according to the International 10-20 system

We used dry electrodes from Florida research instruments (https://fri-fl-shop.com/collections/electrodes-eeg-electrodes). An overview of dry electrodes, their uses, and disadvantages we have described in the paper [10, 2021].

**2. Software**

An overview of software for processing EEG signals was given earlier in this paper [8, 2021]. The main motive for developing our own software rather than integrating it into existing platforms is the expediency of constantly changing and improving the software, for example, for mind control of the robot. We have made the connection of RaspberryPi and ADC in C language and created a static library and implemented the subsequent signal processing in Python. The static library sets the signaling rate and registers for setting up the ADC. For us, it is essential that when a Python file is corrected, the library does not need to be recompiled. Figure 2 shows the process of measuring EEG signals and blink artifacts.

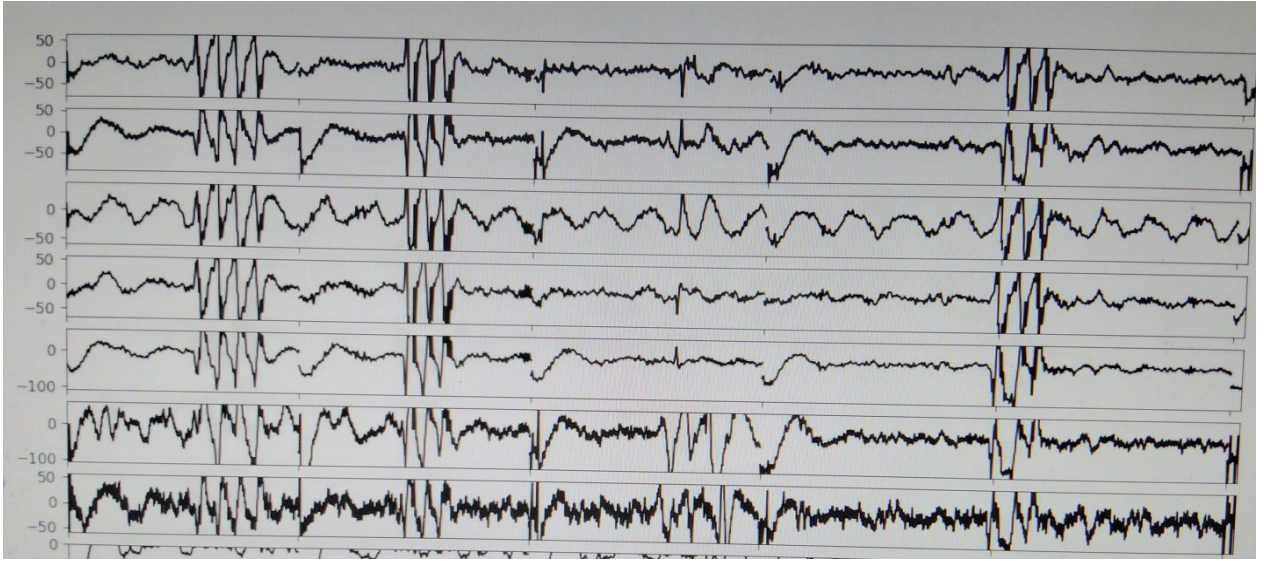

Fig.2. Blinking with EEG signal on PIEEG (real-time)

As shown in the figure, the response in the EEG signals to the blink is available on all channels, so in the research, it is sufficient for us to use only one electrode; we chose the electrode in position - Fz.

**3.1 Device setup**
The position of the electrode can be changed and adjusted for both chewing detection and blinking. For the measurement process, the PIEEG should be connected to the Raspberry PI3 or RaspberryPI4. The power supply for the RaspberryPi must be connected to a battery - 5V (**do not use the power network**).
The method we chose for feature selection is the Fast Fourier Transform, where the signal is decomposed into frequencies [9, 2021]. Repetitive actions such as blinking or chewing. It is worth noting that blinking at a specific frequency is quite tricky, which is done intuitively. The signal amplitude is different for each person (the response of EEG signals to blinking), so the settings for the frequency range of the response and the amplitude of the signal are found by each user through experimental methods. In Figs. 3 and 4, we show the results of controlling two LEDs by blinking. We found amplitude peaks in the frequency ranges of 5 Hz and 3 Hz for the EEG signal decomposed by Fast Fourier Transform.

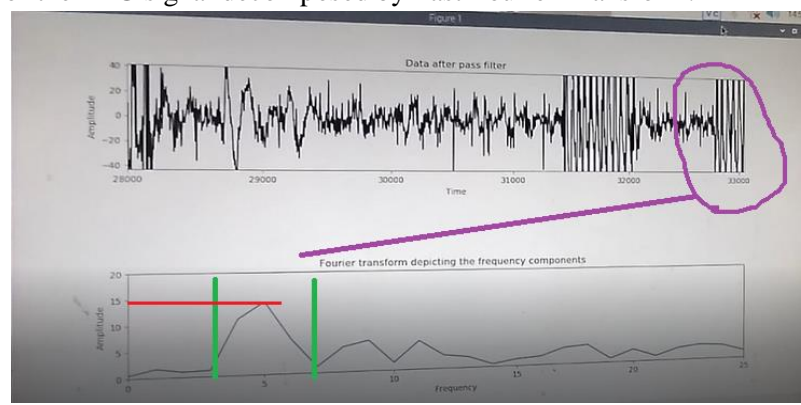

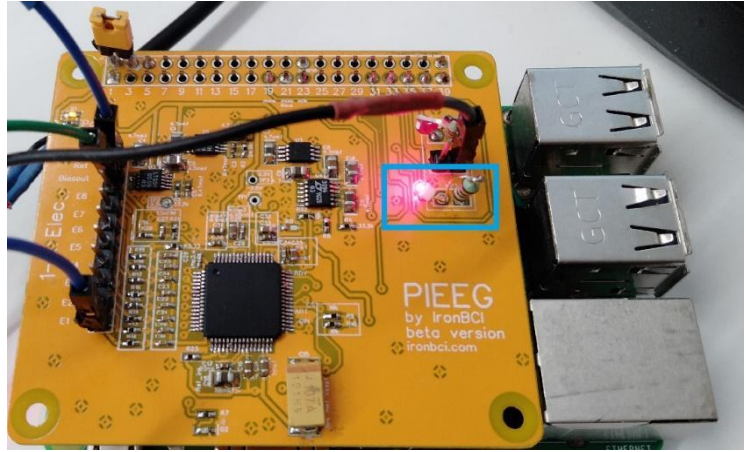

a

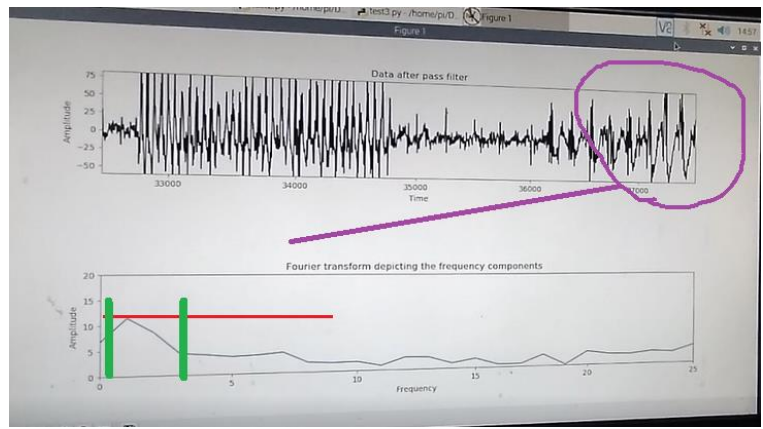

b

Fig.3. LED control in the range of 3 - 7 Hz. a (upper graph) - EEG signal through pass filter 1-30 Hz. an (upper c) – EEG signal (the part circled in purple from the graph above) received through the Fast Fourier Transform. The red line is the setting for turning on the LED, green lines are the frequency range in which we monitor the signal amplitude. b – LED circled in blue, output - 31 for GPIO.setmode (GPIO.BOARD)

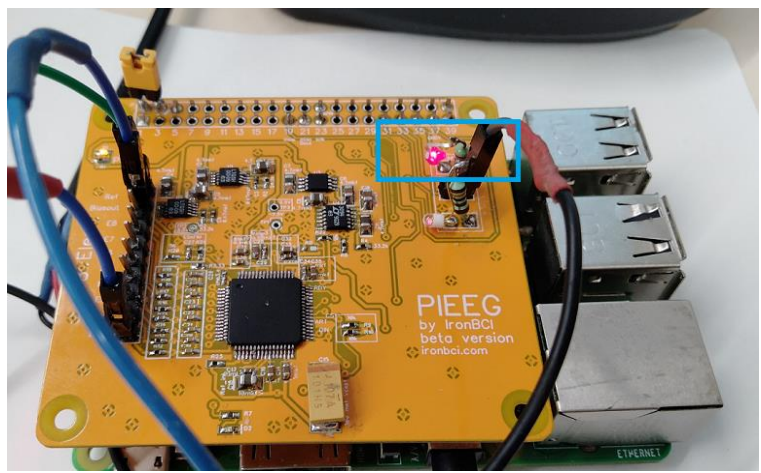

a

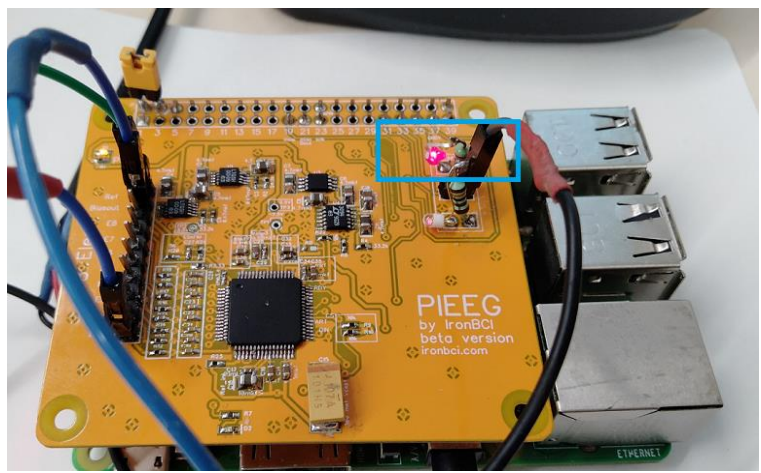

b

Fig.4. LED control in the range of 1 - 3 Hz. a (upper graph) - EEG signal through pass filter 1-30 Hz. a (lower upper graph) – EEG signal (the part circled in purple from the graph above) received through the Fast Fourier Transform. Redline is the setting for turning on the LED, green lines are the frequency range in which we monitor the signal amplitude. b – LED circled in blue, output - 35 for GPIO.setmode (GPIO.BOARD)

The result is positive but does not look ideal. The accuracy can be improved by setting other ranges for the bandpass filter, changing the location of the electrode, changing ADC registers, et cetera. In our case, the instrument worked correctly in 8 out of 10 cases for both frequency ranges. This error can be caused by the hardware characteristics of the instrument, the software, and human factors. For better illustration, we did not connect the signals to the LED but the remote control of the radio-controlled toy and were able to successfully control the toy's movement (Fig. 5).

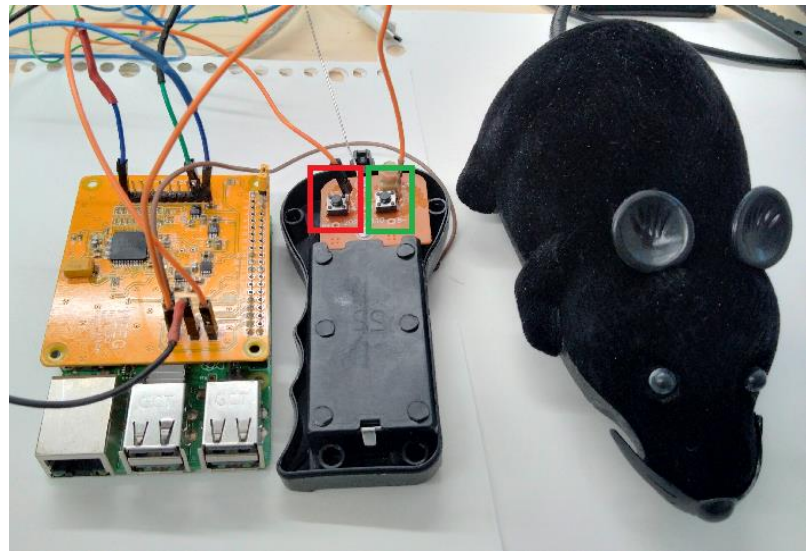

Fig.5. Connecting toys. Discrete signals are connected to the control buttons of the radio remote control. Circled in red and green squares.

The toy is presented only as an example, any type of device can be connected.

**Conclusion**
We suspect that controlling robots by thought would ideally have more to do with motor imagery. However, that has not been done with non-invasive electrodes yet. All the studies we have analyzed generally work in two coordinates - x and y, and the z coordinate. Therefore, the ease of implementation of this method will hopefully inspire engineers to move forward in this direction. We still believe that non-invasive electrodes will soon achieve better results, thanks to the power of machine learning, allowing more natural control of robots.
We did not want to show that blinking is the best way to control a robot, but we wanted to use cheap devices to demonstrate how to control external objects using EEG signals. Paradigms such as P300 and ERPs that allow printing on the keyboard can also be used in similar scenarios.
Now the software works in real time with a delay of 1 second, if necessary, the reaction speed can be increased. One of the advantages of working with a shield is that the data from the ADC is transferred directly to the processor using the SPI protocol. We do not need to use adapters, transmitters, such as Bluetooth, Wi-Fi, Serial. Ideally, when implementing software in a lower programming language, the response speed allows us to receive data at the transfer rate set in the ADS1299 - 250 SPS to 16 kSPS.
The shield can be installed on the OrangePi and BananaPi series SBCs. It is worth noting that the connection to the boards is made through the power supply, SPI channels and several discrete outputs, so the devices can be connected to any SBCs through wires if the GPIO channels do not match. All the software is in the public domain and is under an open license, all rights and comments are welcome. We will continue to work on the software and the shield in the future.